# Spatial and Spectral Quality Evaluation Based On Edges Regions of Satellite Image Fusion

Firouz Abdullah Al-Wassai[1]
Research Student, Computer Science Dept.
(SRTMU), Nanded, India
fairozwaseai@yahoo.com

N.V. Kalyankar[2]
Principal, Yeshwant Mahavidyala College
Nanded, India
drkalyankarnv@yahoo.com

Ali A. Al-Zaky[3]
Assistant Professor, Dept. of Physics, College of Science, Mustansiriyah Un.
Baghdad – Iraq.
dr.alialzuky@yahoo.com

*Abstract*: **The Quality of image fusion is an essential determinant of the value of processing images fusion for many applications. Spatial and spectral qualities are the two important indexes that used to evaluate the quality of any fused image. However, the jury is still out of fused image's benefits if it compared with its original images. In addition, there is a lack of measures for assessing the objective quality of the spatial resolution for the fusion methods. Therefore, an objective quality of the spatial resolution assessment for fusion images is required. Most important details of the image are in edges regions, but most standards of image estimation do not depend upon specifying the edges in the image and measuring their edges. However, they depend upon the general estimation or estimating the uniform region, so this study deals with new method proposed to estimate the spatial resolution by Contrast Statistical Analysis (CSA) depending upon calculating the contrast of the edge, non edge regions and the rate for the edges regions. Specifying the edges in the image is made by using Soble operator with different threshold values. In addition, estimating the color distortion added by image fusion based on Histogram Analysis of the edge brightness values of all RGB-color bands and L-component.**

***Keywords: Spatial Evaluation; Spectral Evaluation; contrast; Signal to Noise Ratio; Measure of image quality; Image Fusion***

## I. INTRODUCTION

Many fusion methods have proposed for fusing high spectral and spatial resolution of satellite images to produce multispectral images having the highest spatial resolution available within the data set. The theoretical spatial resolution of fused images $F$ is supposed to be equal to resolution of high spatial resolution panchromatic image$PAN$; but in reality, it reduced. Quality is an essential determinant of the value of surrogate digital images. Quantitative measures of image quality to yield reliable image quality metrics can be used to assess the degree of degradation. Image quality measurement has become crucial for most image processing applications [1]. With the growth of digital imaging technology over the past years, there were many attempts to develop models or metrics for image quality that incorporate elements of human visual sensitivity [2]. However, there is no current standard and objective definition of spectral and spatial image quality. Image quality must be inferred from measurements of spatial resolution, calibration accuracy, and signal to noise, contrast, bit error rate, sensor stability, and other factors [3]. Most important details of the spatial resolution image are included in edges regions, but most of its standards assessment does not depend upon specifying edges in the image and measuring their edges, but they depend upon the general estimation or estimating the uniform region [4-6].

Therefore, in this study, a new scheme for evaluation, spatial quality of the fused images based on Contrast Statistical Analysis (CSA), and it depends upon the edge and non-edge regions of the image. The edges of the image are made by using Soble operator with different thresholds, and in comparing its results with traditional method of MTF depending upon the uniform region of the image as well as on completely image as the metric evaluation of the spatial resolution. In addition, this study testifies the metric evaluation of the spectral quality of the fused images based on Signal to Noise Ratio SNR of image upon separately uniform regions and comparing its results with other method depends on whole MS & fused images.

The paper is planned in five sections that are as follows: Section I, which is considered the introduction of the study, brings framework and background of the study, Section II illustrates the quality evaluation of fused images i.e., a new proposed scheme of spatial evaluation quality of fused images defined as Contrast Statistical Analysis Technique CSA. Section III brings experimenting and analyzing results of the study based on pixel and feature level fusion including: High –Frequency-Addition Method (HFA)[20], High Frequency Modulation Method (HFM) [7], Regression variable substitution (RVS) [8], Intensity Hue Saturation (IHS) [9], Segment Fusion (SF), Principal Component Analysis based Feature Fusion (PCA) and Edge Fusion (EF) [10]. All these methods will mention in section IV. Section V will be the conclusion of the study.

## *II. QUALITY EVALUATION OF THE FUSED IMAGES*

The quality Evaluation of the fused images clarified through describing of various spatial and spectral quality metrics that used to evaluate them. With respect to the original multispectral images MS, the spectral fidelity of the fused images is described. The spectral quality of the fused images analyzed by compare them with spectral characteristics of resampled original multispectral images$M_k$. Since the goal is to preserve the radiometry of the original MS images, any metric used must measure the amount of change in digital number values of the pan-sharpened or fused image $F_k$ and compared to the original image$M_k$ for each of band k. In order to evaluate the spatial properties of the fused images, a panchromatic image PAN and intensity image of the fused image have to be compared since the goal is to retain the high spatial resolution of the PAN image.

### *A. The MTF Analysis*

This technique defined as Modulation transfer function (MTF)[3] and referred to Michelson Contrast $C_M$ . In order to calculate the spatial resolution by this method, it is common to measure the contrast of the targets and their background [11]. In this study, I used this technique in equation (1) to calculating the contrast rating based on uniform regions as well as overall images. The homogenous regions selected (see Fig. 11) have the size as the following:

1. $30 \times 30$ Block size for two different homogenous regions named b1 b2 respectively.
2. $10 \times 10$ Block size for seven different homogenous regions at same time named b3.

Contrast performance over a spatial frequency range is characterized by the $C_M$ [3]:

$$C_M = \frac{I_{max} - I_{min}}{I_{max} + I_{min}} \quad (1)$$

Where $I_{max}$ $I_{min}$ are the maximum and minimum radiance values recorded over the region of the homogenous image. For a nearly homogeneous image, $C_M$ would have a value close to zero while the maximum value of $C_M$ is 1.0.

### *B. Signal-to Noise Ratio (SNR)*

The signal-to-noise ratio SNR is a measure of the purity of a signal [11]. Other means measuring the ratio between information and noise of the fused image [12]. Therefore, estimation of noise contained within image is essential which leads to a value indicative of image quality of the spectral resolution. Here, this study proposes to estimate the SNR based on regions for evaluation of the spectral quality. Also, results of the SNR based on regions that was compared with other results of the SNR based on whole MS and Fused images employed in our previous studies [13]. The two methods as the following:

1. *$SNR_a$ Based On Regions*

The SNR evaluation is Similar to contrast analysis technique, the final SNR rating is based on a $30 \times 30$ block size for two different regions of the homogenous as well as seven different regions at same time a 10x10 block size (see fig.3) image calculation of all RGB-color bands $k$. Which reflects the SNR across the whole image, the SRN in this implementation defined as follows [14]:

$$SNR_{a_k} = \frac{\mu_k}{\sigma_k} \quad (2)$$

Where: $SNR_{a_k}$ Signal-to Noise Ratio, $\sigma$ standard deviation and $\mu$ the mean of brightness values of RGB band $k$ in the image region. The mean value $\mu_k$ is defined as [15]:

$$\mu_k = \frac{1}{mn} \sum_{i=1}^{n} \sum_{j=1}^{m} f_k(i,j) \quad (3)$$

The standard deviation σ is the square root of the variance. The variance of image reflects the dispersion degree between the brightness values and the brightness mean value. The larger σ is more disperse than the gray level. The definition of σ is [15]:

$$\sigma_k = \sqrt{\frac{1}{mn}} \sum_{i=1}^{n} \sum_{j=1}^{m} (f_k(i,j) - \mu_k)^2 \quad (4)$$

2. *$SNR_b$ Based On Whole MS With Fused Images*

In this method, the signal is the information content of original MS image$M_k$, while the merging $F_k$ can cause the noise, as error that is added to the image fusion. The signal-to-noise ratio$SNR_k$, given by [16]:

$$SNR_{b_k} = \sqrt{\frac{\sum_i^n \sum_j^m (F_k(i,j))^2}{\sum_i^n \sum_j^m (F_k(i,j) - M_k(i,j))^2}} \quad (5)$$

The SNR therefore is a relative value that reflects the percentage of significant values representing borders of objects. Thus, the SNR can be used to generate an indication of image quality of spectral resolution in dependence on the results of analyzing the image data. In the first method the result of $SNR_a$ should has highly dissimilar to the results of MS as possible. In the second method, the maximum value of $SNR_b$ is the best image to preservation of the spectral quality for the original MS image.

### *C. The Histogram Analysis*

The histograms of the multispectral original MS and the fused bands must be evaluated [17]. If the spectral information preserved in the fused image, its histogram will closely resemble the histogram of the MS image. The analysis of histogram deals with the brightness value histograms of all RGB-color bands, and L-component of the resample MS image and the fused image that computed the edges of image's points regions only by using the next technique to estimated the edge regions. A greater difference of the shape of the corresponding histograms represents a greater spectral change [18].

## *III. CSA a New Scheme Of Spatial Evaluation Quality of The Fused Images*

To explain the new proposed technique of Contrast Statistical Analysis CSA for evaluation the quality of the spatial resolution specifying the edges in the image by using Soble operator. In this technique the metric starts by applying Soble edge detector for the whole image [19, 20], but the new proposed method based on contrast calculation of each of the edge and homogenous regions. The steps for evaluation of the spatial resolution as follows:

1- Apply Soble edge detector for the whole image with different thresholds of its operator i.e. 20, 40, 60, 80 and 100.

2- The pixel value of the image is labeled into edge regions or homogenous regions in corresponding with applied Soble thresholds. If the pixel number value is greater than a certain predefined threshold, it (the pixel) is labeled as an edge point, otherwise, it is considered smooth or homogenous region and further processing is disabled.

3- Calculated the rate of the strong edges pixels for all RGB bands $k$ with different thresholds of Soble operators and drawing the histograms for them as well.

4- Estimate the mean $\mu$ and standard deviation $\sigma$ for all RGB bands $k$ of all edges points and homogenous regions.

5- Finally, CSA was calculated by the statistical characteristics of the edges points and homogenous regions for all RGB color bands $k$ in image were adopted according to equation (1). Here, the $I_{min}$ & $I_{max}$ are calculated by adopting the mean $\mu$ (eq.3)and standard deviation $\sigma$ (eq.4) of edges regions at (n, m) for the intensity $f_k(i,j)$ of image components relating to points and homogenous regions according to the two following relations:

$$I_{k_{min}} = \mu_k - \sigma_k \quad \& \quad I_{k_{max}} = \mu_k + \sigma_k \quad (6)$$

$$\therefore CSA_k = \frac{\sigma_k}{\mu_k} \quad (7)$$

Where: CSA contrast of band $k$, µ mean, $\sigma_k$ standard deviation. For a nearly homogeneous image, $CSA_k$ would have a value close to zero while the maximum value of $CSA_k$ is 1.0. The maximum contrast value for the image means that it has the high spatial resolution.

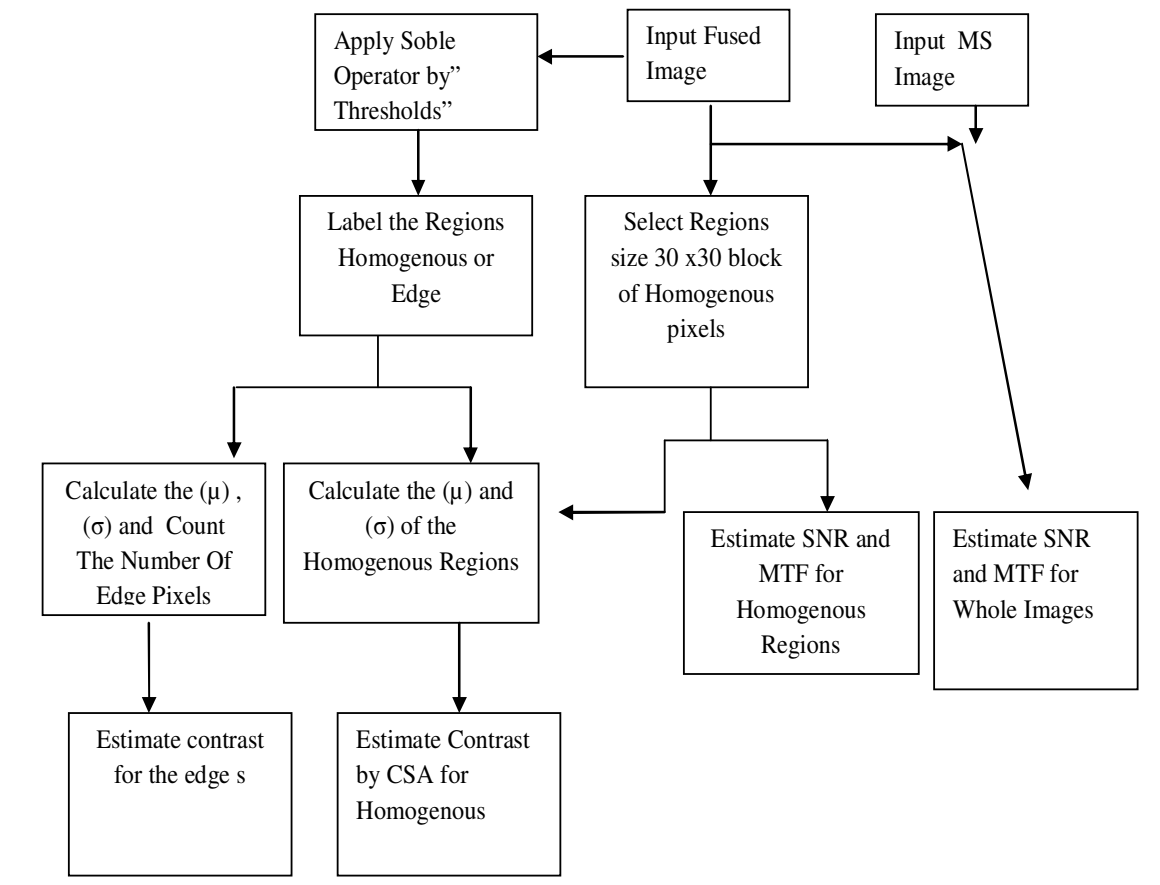

Fig.1: Schematic Flowchart of Spatial and Spectral Evaluation Quality Image Fusion

## IV. EXPERIMENTAL &ANALYSIS RESULTS

The above assessment techniques are tested on fusion of Indian IRS-1C PAN (0.50 - 0.75 µm) of the 5.8 m resolution panchromatic PAN band and the Landsat TM red (0.63 - 0.69 µm), green (0.52 - 0.60 µm) and blue (0.45 - 0.52 µm) bands of 30 m resolution multispectral image MS were used in this work. Fig.2 shows IRS-1C PAN and multispectral MS TM images. Hence, this work is an attempt to study the quality of the images fused from different sensors with various characteristics. The size of the PAN is 600 * 525 pixels at 6 bits per pixel and the size of the original multispectral is 120 * 105 pixels at 8 bits per pixel, but this is upsampled by nearest neighbor. The pairs of images were geometrically registered to each other. Fig. 2 shows the fused images of the HFA, HFM, HIS, RVS, PCA, EF, and SF methods are employed to fuse IRS-C PAN and TM multi-spectral images. To simplify the comparison of the different fusion methods, the results of the fused images are provided as charts from Fig. 3 to Fig.12 for quantify the behavior of HFA, HFM, IHS, RVS, PCA, EF, and SF methods.

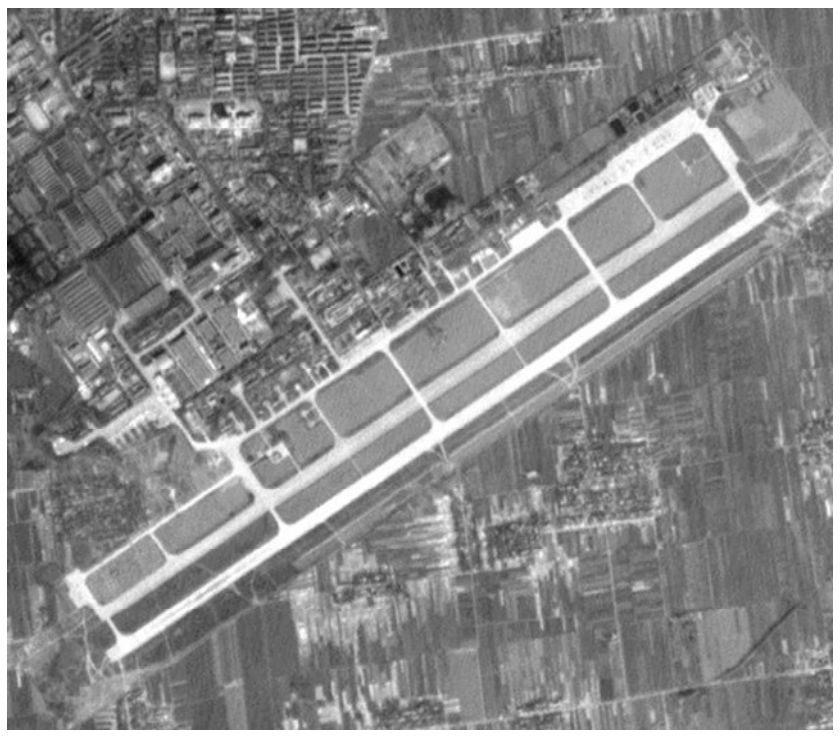

Original PAN Image

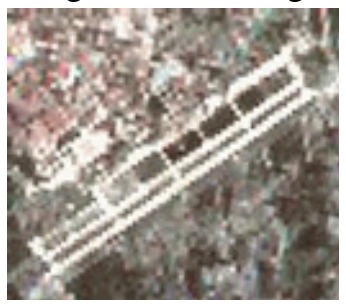

Original MS Image

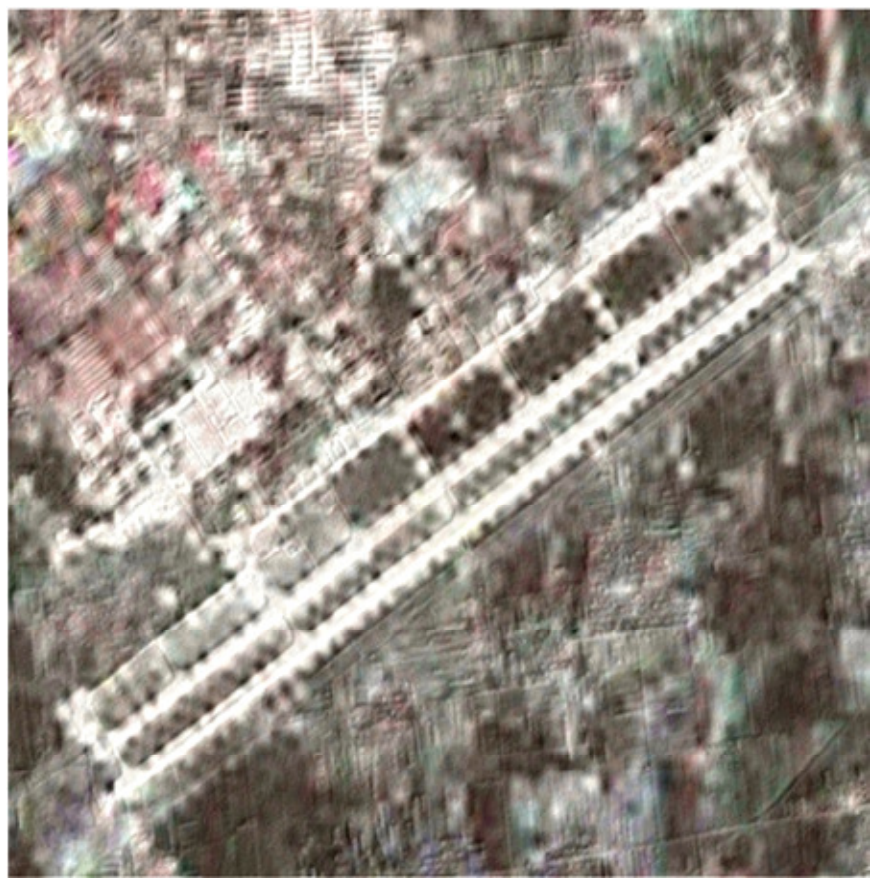

HFA

Fig.1: The Representation of Original & Fused Images

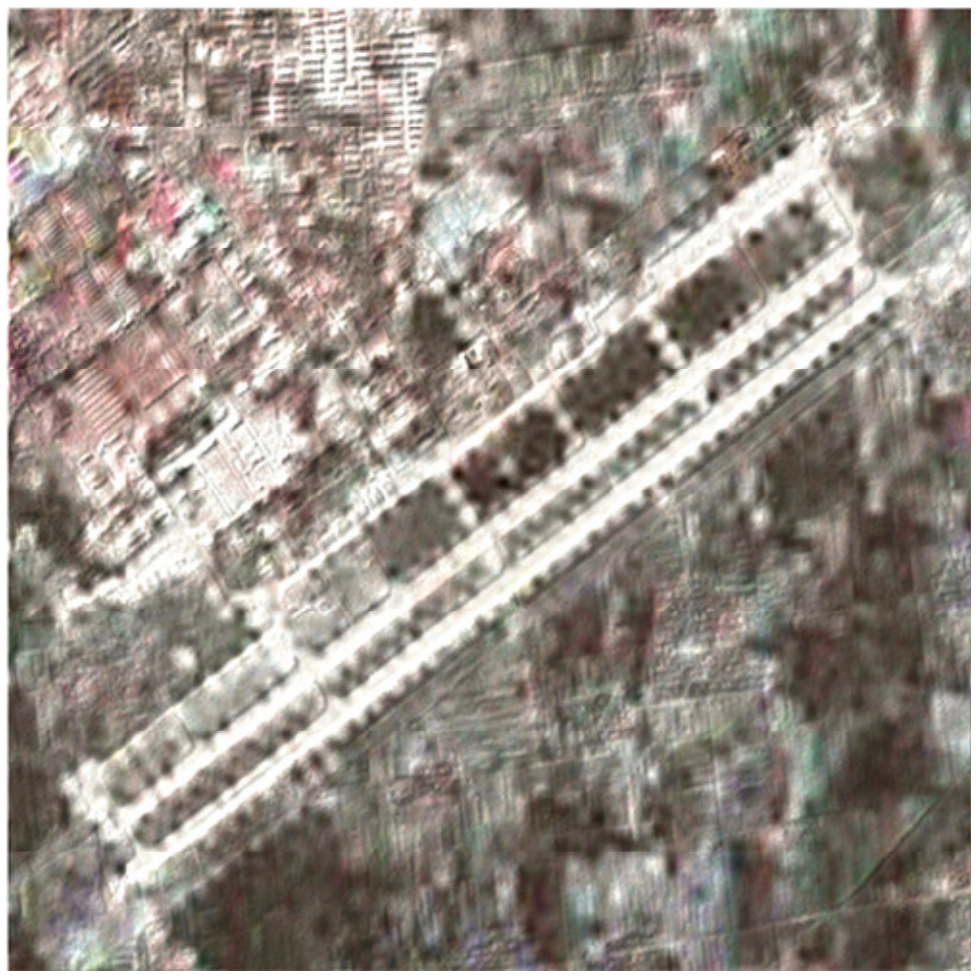

HFM

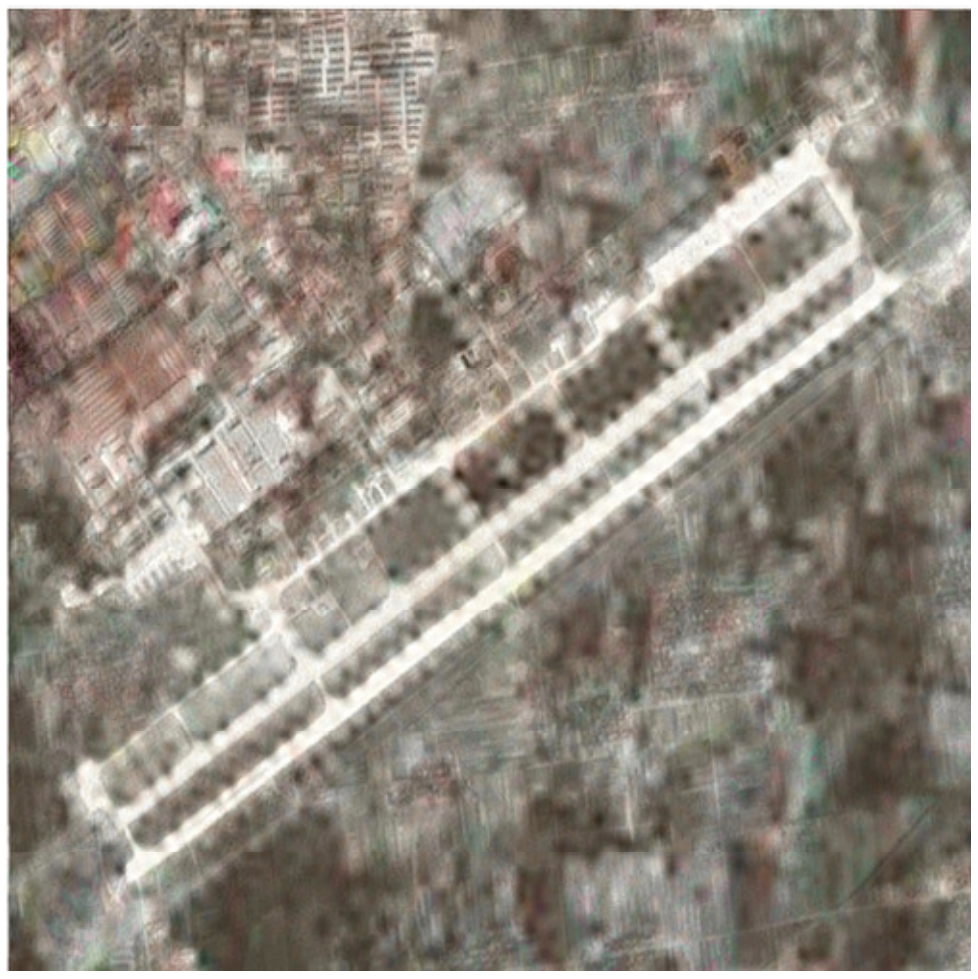

IHS

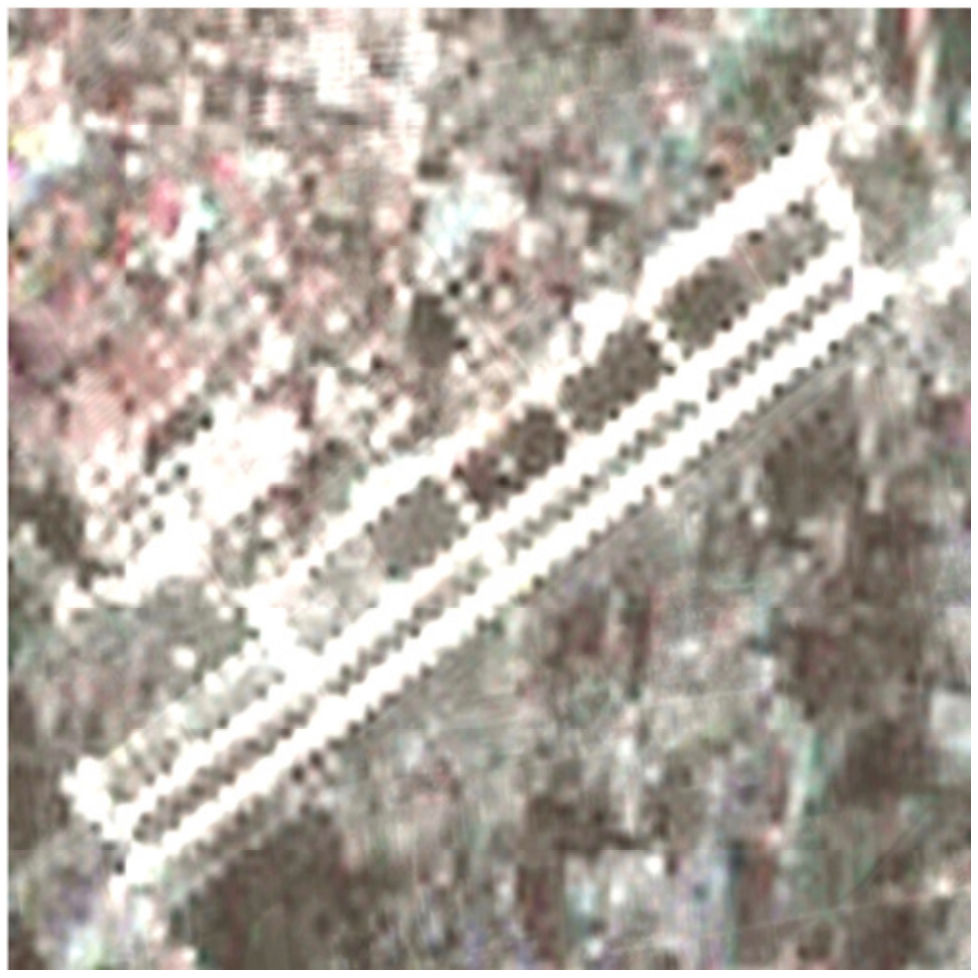

PCA

Continue Fig.2: The Representation of Original & Fused Images

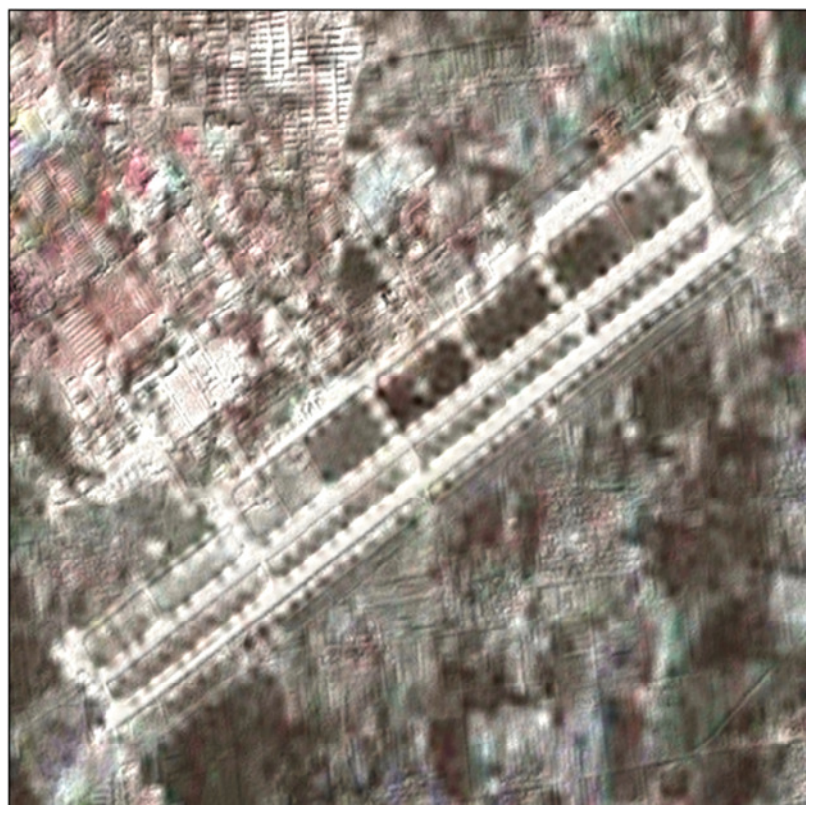

RVS

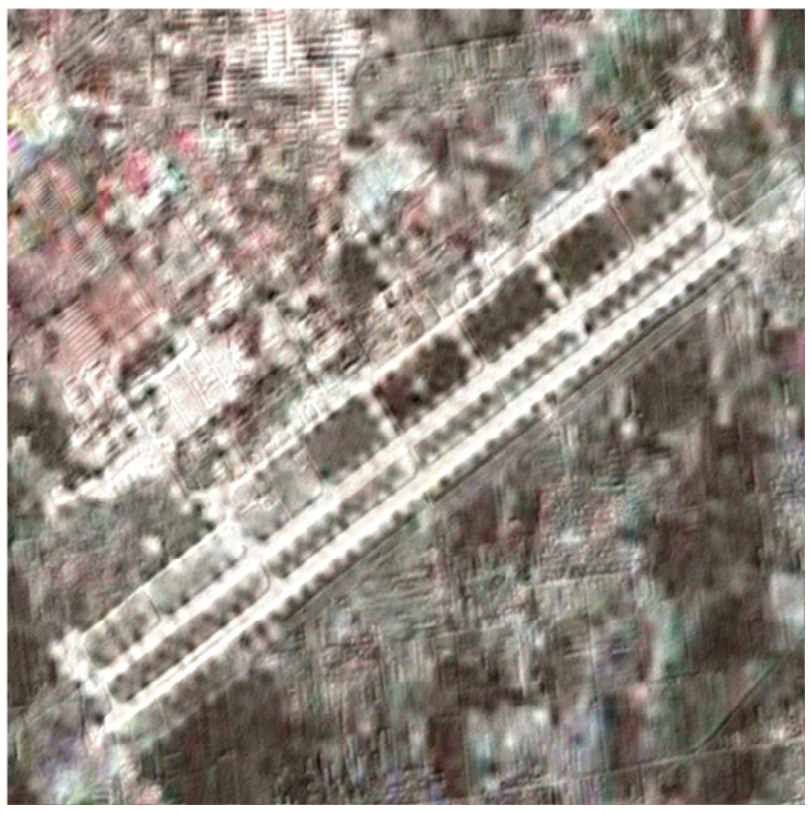

SF

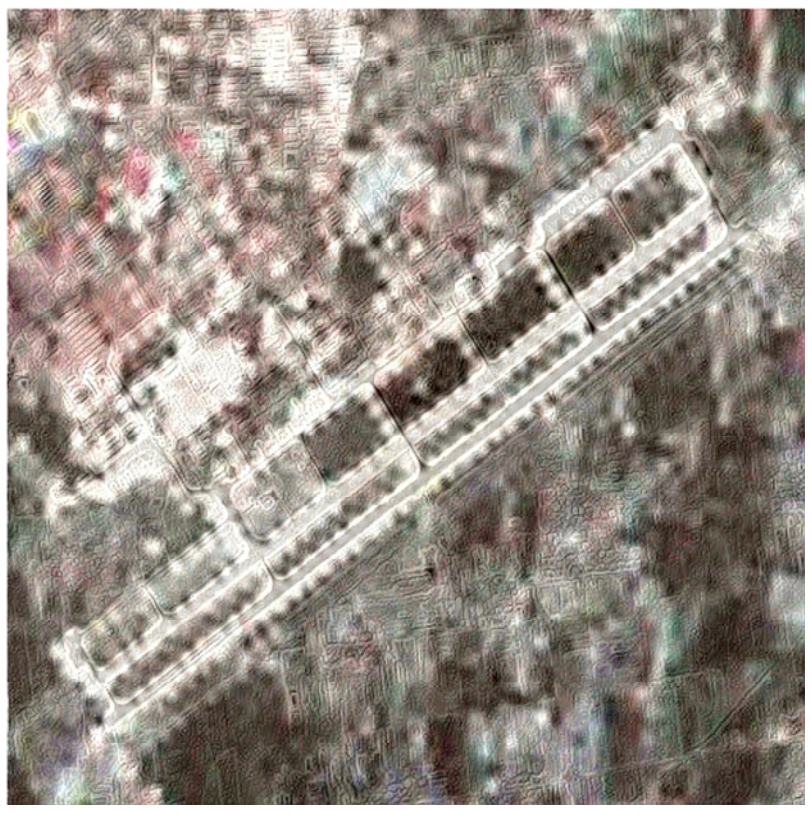

EF

Continue Fig.2: The Representation of Original & Fused Images

### A. *Spatial Quality Metrics Results*

From Fig. 3 and Fig.4, it is clear that the differences results between MTF and CSA techniques depending on the whole images. The MTF gives the same results that shown high contrast for all methods of the EF, HFA and RVS methods. While the results of CAS different results with results MTF for all methods. And also, when comparing the results of MTF with CAS based on the homogenous certain regions (see the certain homogenous regions in Fig. 7), MTF gives same results of the bands (G & B) approximately for the specific homogenous regions b2 and b3 of image fusion methods. It is obvious that the result of CSA is better than MTF since the CSA gave the smallest different ratio between the image fusion methods. Generally, According to the computation results, CSA based on whole regions in Fig.4 & Fig.6 and the maximum contrast was for EF methods where the other methods that have high contrast than the original of MS image except IHS and PCA methods. The EF method has many details of information however; it is appearing not really information as the PAN image because this technique depending on the sharpening filters.

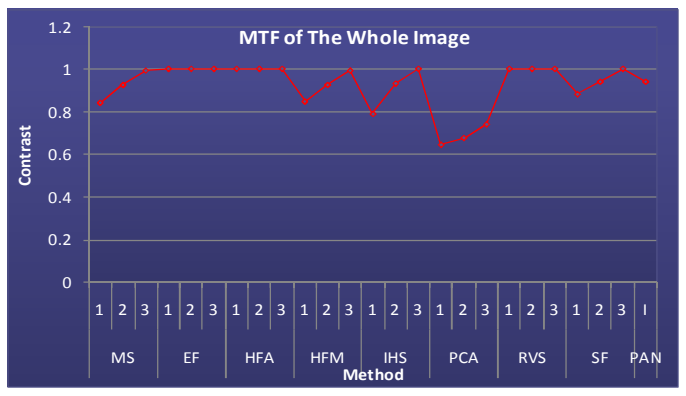

Fig.3: The MTF Analysis Technique for whole of the Image Fusion Methods

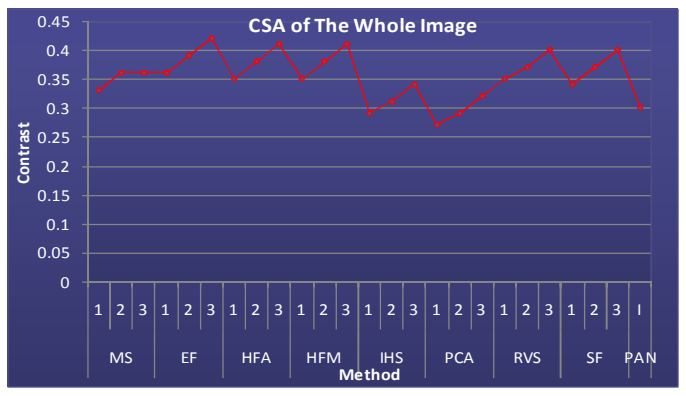

Fig.4: CSA Technique for Whole of the Image Fusion Methods

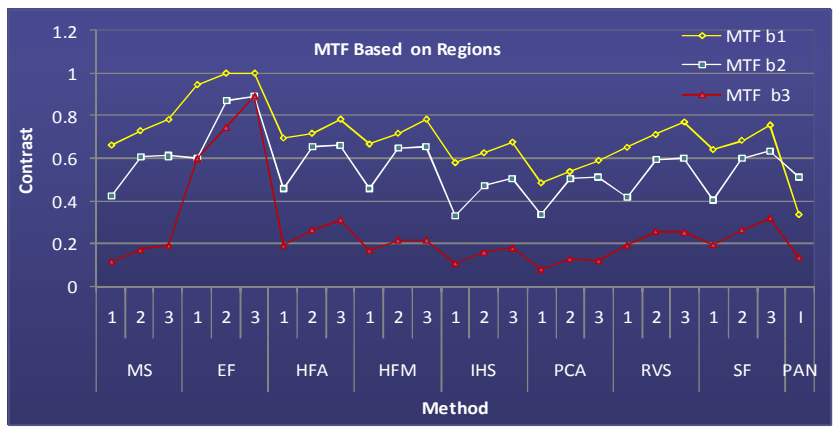

Fig.5: The MTF Analysis Technique for Selected Homogenous Region of the Image Fusion Methods

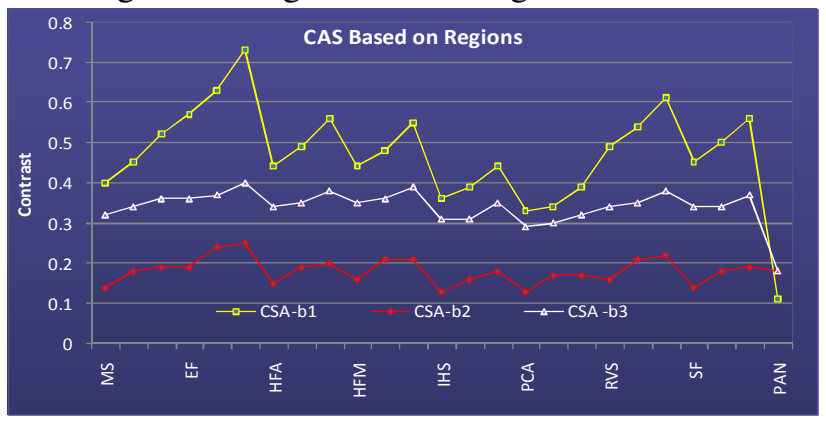

Fig.6: CSA Technique for Selected Homogenous Region of the Image Fusion Methods

Fig.8 shows the results of CSA proposed method. It is evident that of this metric provides the accurate results with each band in Fig.8 are better than previous criteria that based on region or completely image. Because of CSA, the criteria that approved on the edge by Soble operator do not subject to choice the homogenous region that may possibly not be the same in Fig.5 & Fig.6. For instance the results of the homogeneous been selected were the results of Fig.6 are different despite using the same criteria of the CSA. It is important to observe that the results increase the contrast of the merged image more than the original MS image and at the same time it should be same or nearby the results of the PAN image. According to the computation results, CSA in Fig.8 based on the edge regions by Soble operator with different values of the thresholds, it is clearly that the CSA results of fused images improved the spatial resolution for all methods especially the maximum values with EF technique except that of IHS & PCA methods which obtained on the lower results. It is very appears in Fig.9 and emphasizes those results by calculating the rate of the edge points. When comparing the results of CSA of the edge for the fusion images with edge results of the PAN image in Fig.8, we found that SF, HFA & RVS were the closest to the PAN image results. These methods are better than the highest result which obtained by CSA, and this means the fusion results for the SF, HFA and RVS methods have kept most of spatial information in the original PAN image.

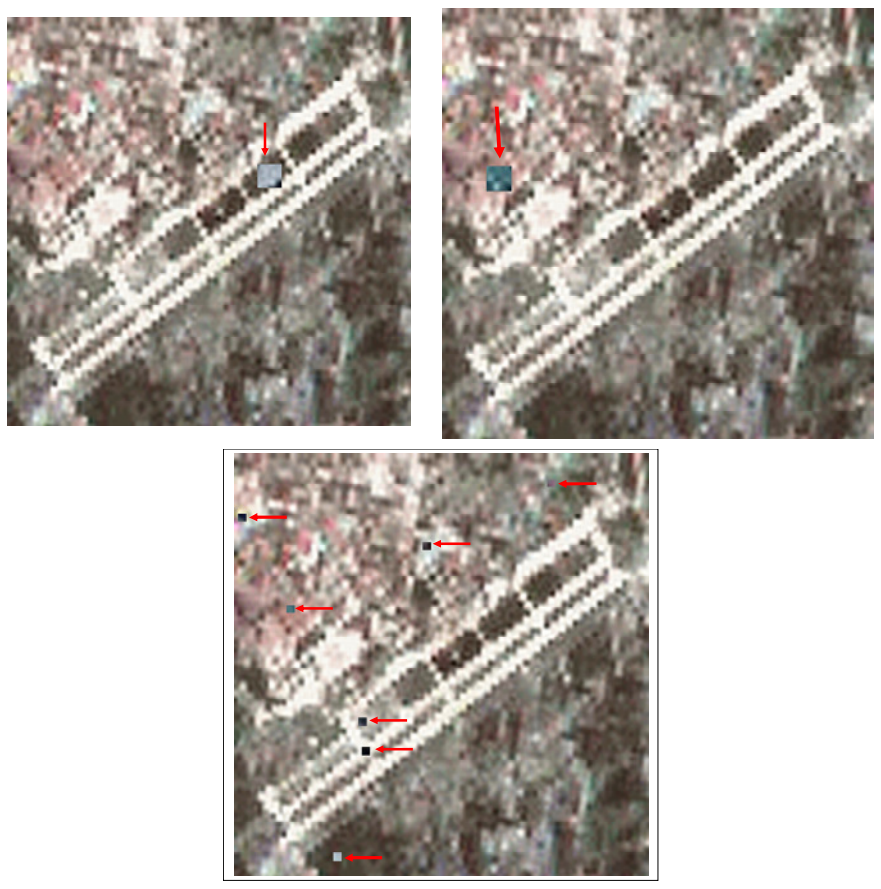

Fig 7: the selected Homogenous Regions as following (a) with 30x30 B1 ,(b) 30x30 B2 and (c) 10x10 B3 for seven homogenous Regions

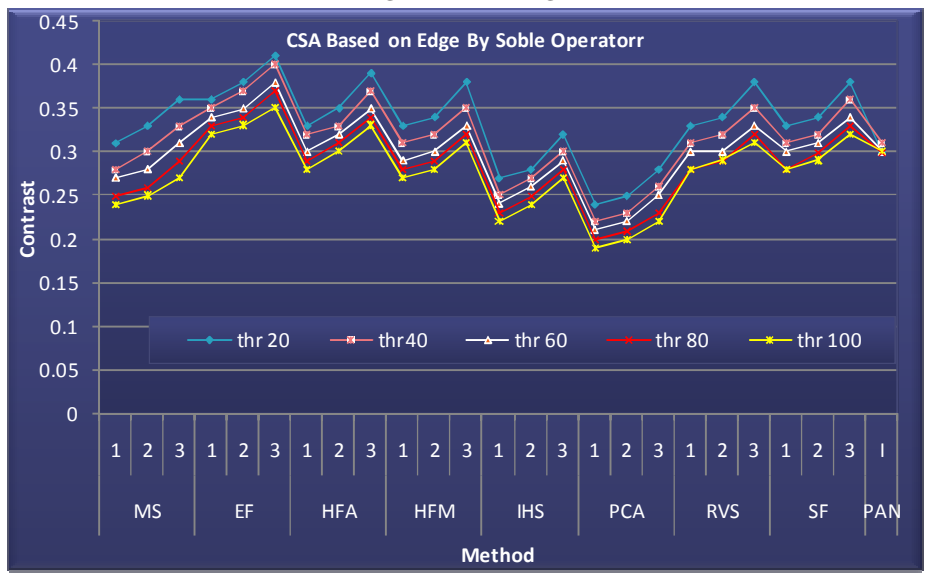

Fig.8: CSA Based On Edge Regions By Soble Operator With Different Threshold

By analyzing impact change of the threshold values on CSA results in Fig.8, it observed that the number of edges decreased when the threshold values increasing as a relationship inverse. However, it appears in Figure 9 not affected by the values of CSA that based on homogeneous regions according to threshold values change as observed in previous results of the edges in Fig.8. It can be absorbed the effectiveness of the improvement spatial for the merging used CSA through homogeneous regions by Soble operator in Fig.10 that does not appear the difference accurately. Despite applied the same of threshold values as applied on the edges image in Fig.8. Because that the edges are really showing the improvement of the spatial resolution of the images, while not appear that the spatial improvement in the homogeneous regions.

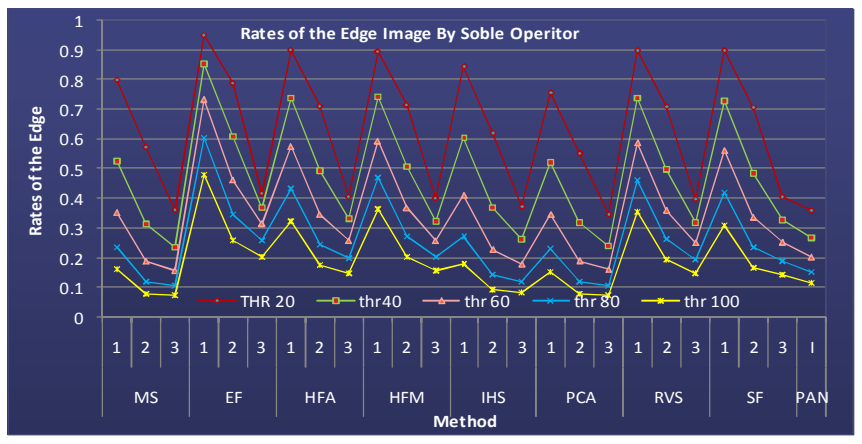

Fig.9: Image Edge Rates Measure by Soble Operator

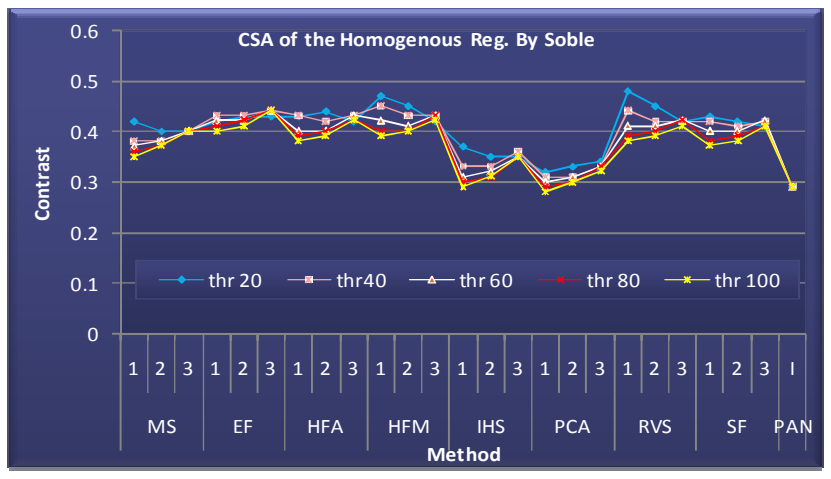

Fig.10: CAS Based On Homogenous Regions by Soble Operator with Different Threshold

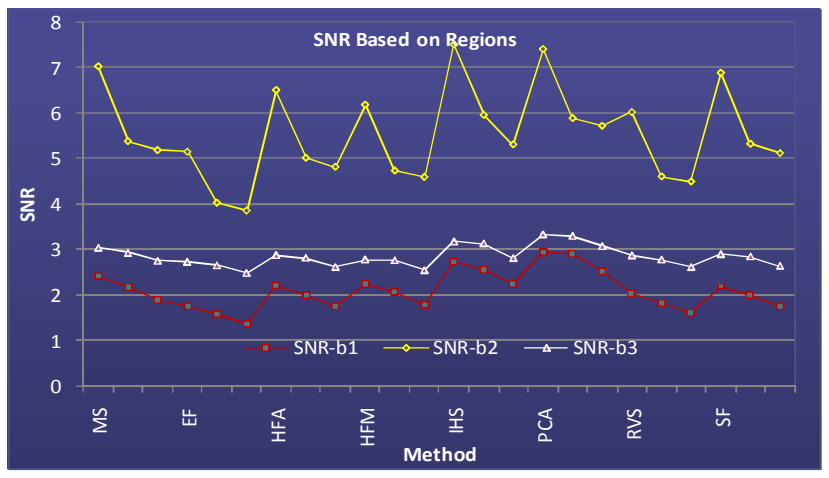

Fig.11: $SNR_a$ Based on homogenous of Regions Image

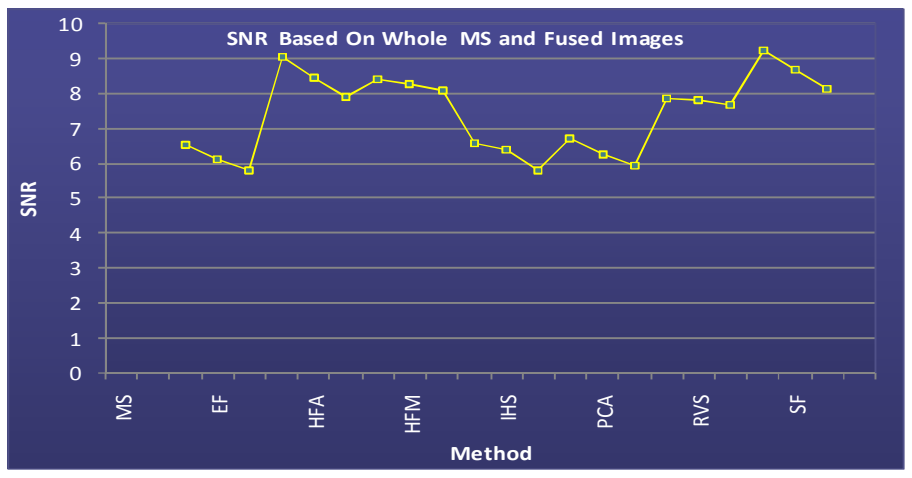

Fig. 12: $SNR_b$ Based on Whole Of The Image

## B. *Spectral Quality Metrics Results*

Using two different measuring evaluation techniques are the SNR and Histogram analysis to testify the degree of color distortion caused by the different fusion methods as the following:

The analyzing SNR of the spectral quality for the image fusion methods based on the regions that using eq.2, the results shown in Fig. 11. It is clearly SNR has different results for each homogenous region. This means SNR has various results dependence on the selected region. It is obvious from the results of the SNR in Fig.11 for example SF has best results followed by HFA method in the region b1, the same results of original MS image, but in other regions the results were closely. Analyzing the results of SNR based on whole images used eq.5 in Fig.12. According to that computation of SNR result in Fig.12, the maximum values were with SF& HFA methods where the lowest values for the IHS and PCA methods. That means the SF & HFA methods preserve the maximum possible to the spectral quality of the original *MS* image. The spectral distortion introduced by the fusion can be analyzed the histogram for all RGB color bands and L-component that based on edge region by Soble with threshold value 20 appears changing significantly. Fig.13 noted that the matching for R &G color bands between the original MS with the fused images. Many of the image fusion methods examined in this study and the best matching for the intensity values between the original MS image and the fused image for each of the R&G color bands obtained by SF. There are also matching for the B color band in Fig.13 and L-component in Fig.14 except when the values of intensity that ranging in value 253 to 255 not appear the values intensity of the original image whereas highlight the values of intensity of the merged images clearly in the Fig.13 & 14. That does not mean its conflicting values or the spectral resolution if we know that the PAN band (0.50 - 0.75 µm) does not spectrally overlap the blue band of the MS (0.45 - 0.52 µm).

Means that during the process of merging been added intensity values found in the PAN image and there have been no in the original MS image which are subject to short wavelengths affected by many factors during the transfer and There can be no to talk about these factors in this context. Most researchers histogram match the PAN band to each MS band before merging them and substituting the high frequency coefficients of the PAN image in place of the MS image's coefficients such as HIS &PCA methods . However, they have been found where the radiometric normalization as IHS &PCA methods is left out Fig. 13, 14, 15 &16.

Generally the best methods through the previous analysis of the Fig.13 and Fig.14 to preservation of the maximum spectral characteristics as possible to the original image for each RGB band and L-component was with SF method where SF results has given of matching with the values of the intensity of original MS.

By analyzing the histogram of the Fig. 15 for the whole image, we found that the values of intensity are less significantly when values of 255 for the G &B-color bands of the original MS image. The extremism in the Fig. 16 for the intensity of luminosity disappeared. The comparison between the results of the histogram analyze for the intensity values at the whole image with the previous results of the Fig. 13 &14 are based on the values of the intensity of the edges. We found that the analysis of spectral distortions by using accurate analysis of the edge for the whole image confirms that the conclusion is the results in the Fig.14&16 of luminosity. The edges are affected more than homogenous regions through the process of merge by moving spatial details to the multispectral MS image and consequently affect on its features and that showed in the image after the merged. Moreover, the best results of the histogram analysis in Fig.15 &16 obtained by the SF technique.

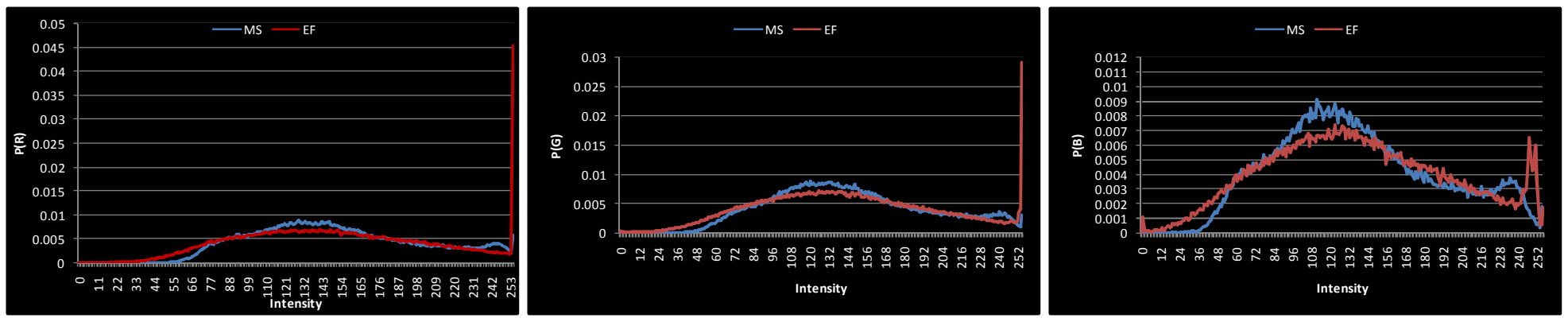

EF

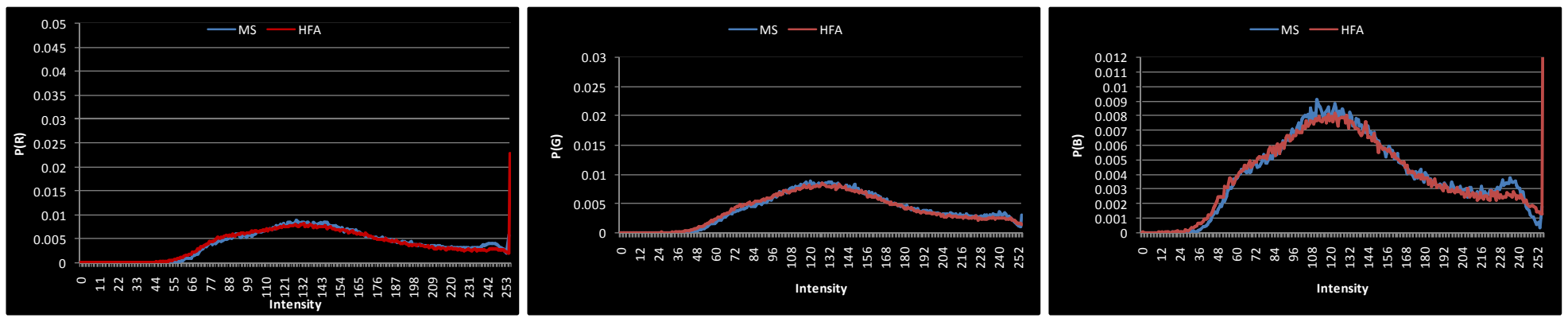

HFA

Fig.13 : Histogram Analysis for All (RGB) Color Bands of Edge of Sharpen Images with Edge Of MS image by Soble Operator with 20 thresholds for the Image Fusions and MS Image

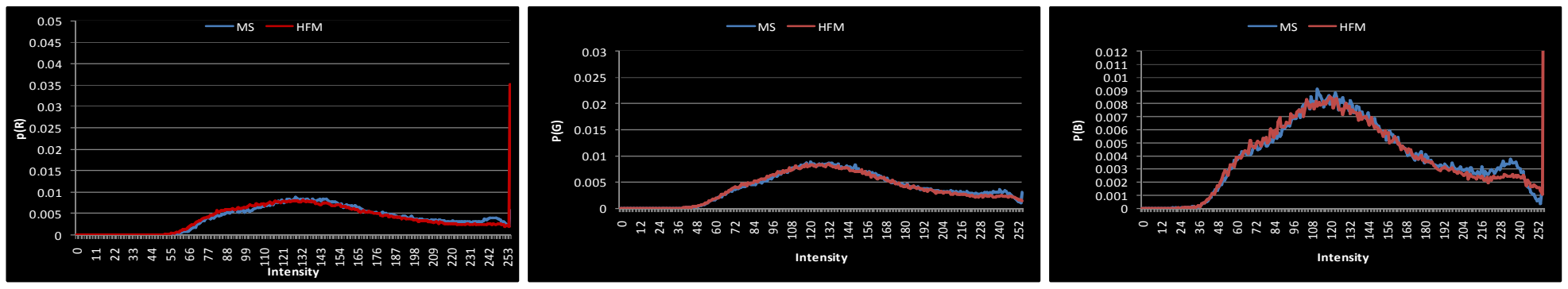

HFM

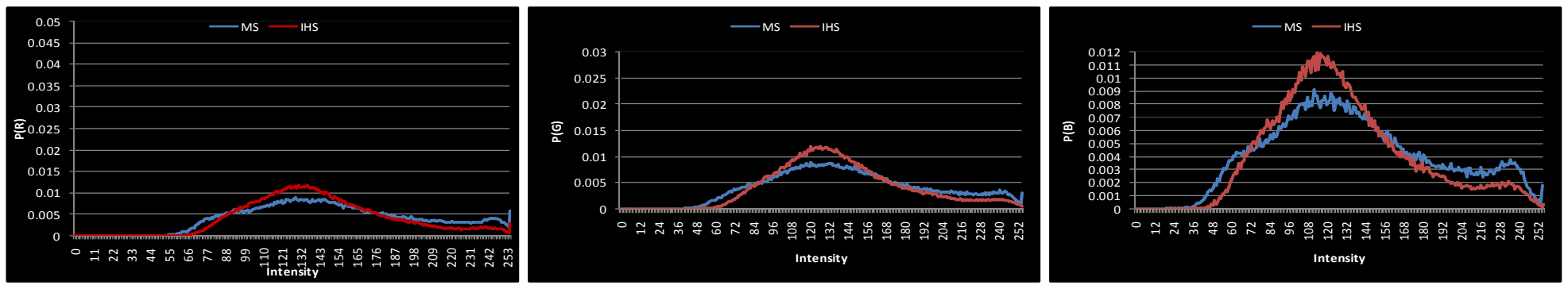

IHS

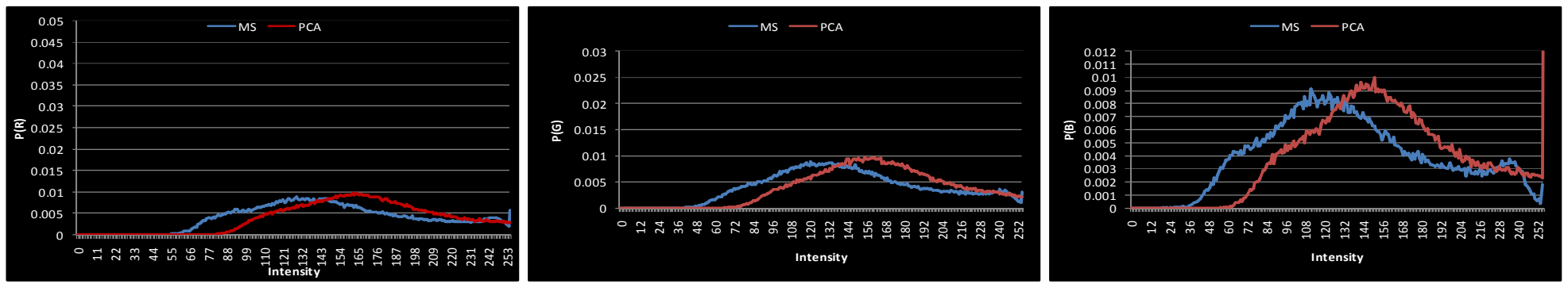

PCA

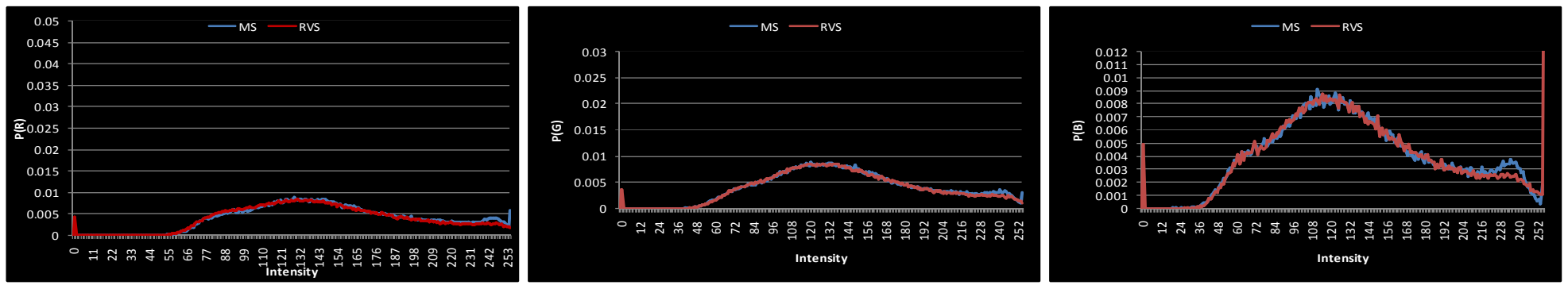

RVS

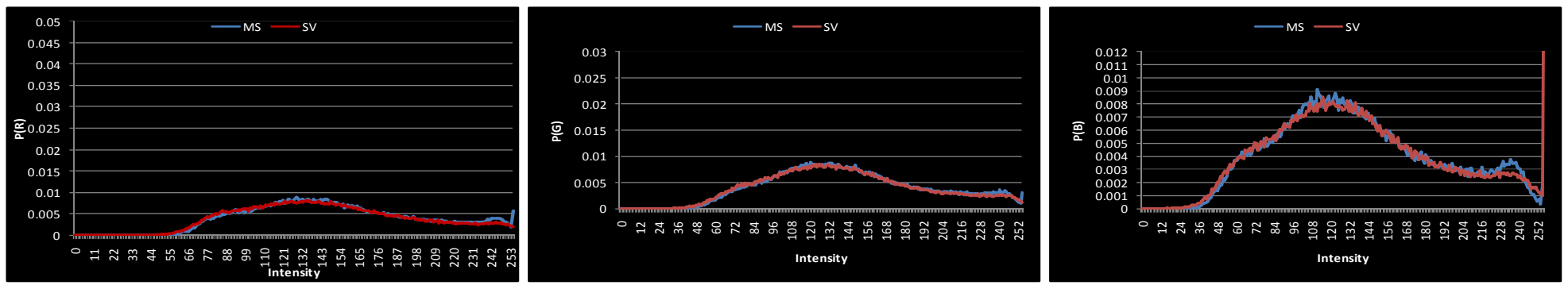

SF

**Continue** Fig.13 : Histogram Analysis for All (RGB) Color Bands of Edge of Sharpen Images with Edge Of MS image by Soble Operator with 20 thresholds for the Image Fusions and MS Image

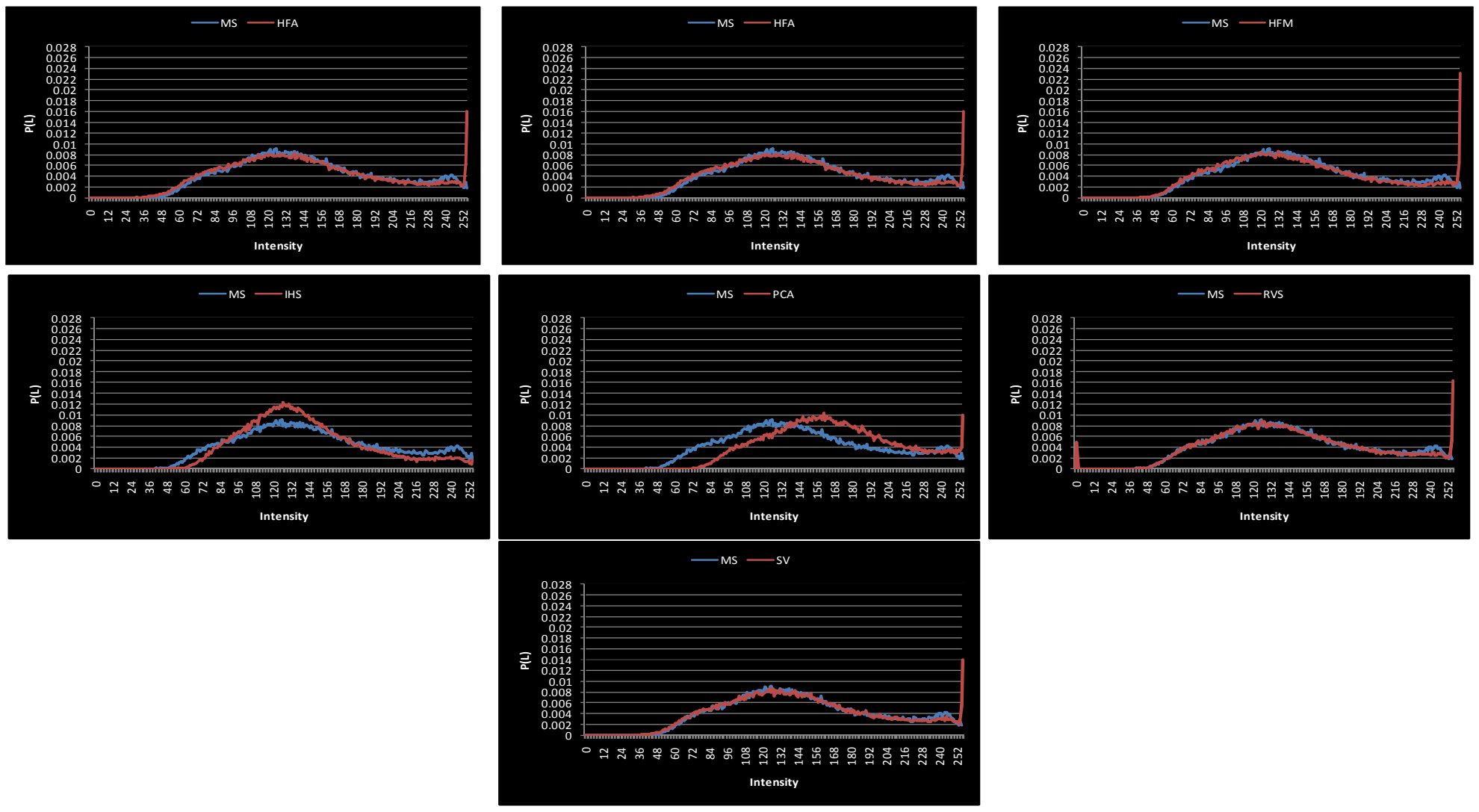

**Fig.14**: Histogram Analysis L-component of Edge Fused Images with Edge of MS image by Soble Operator with 20 thresholds

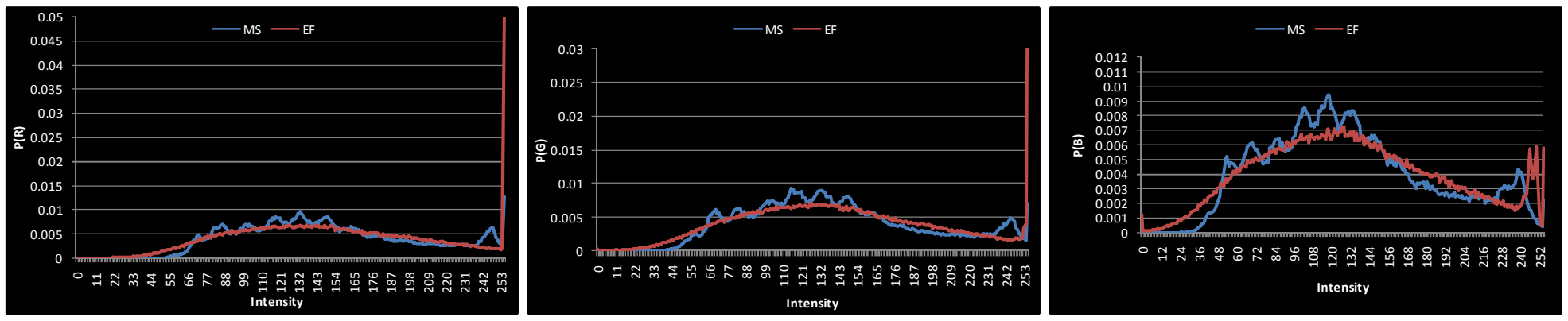

EF

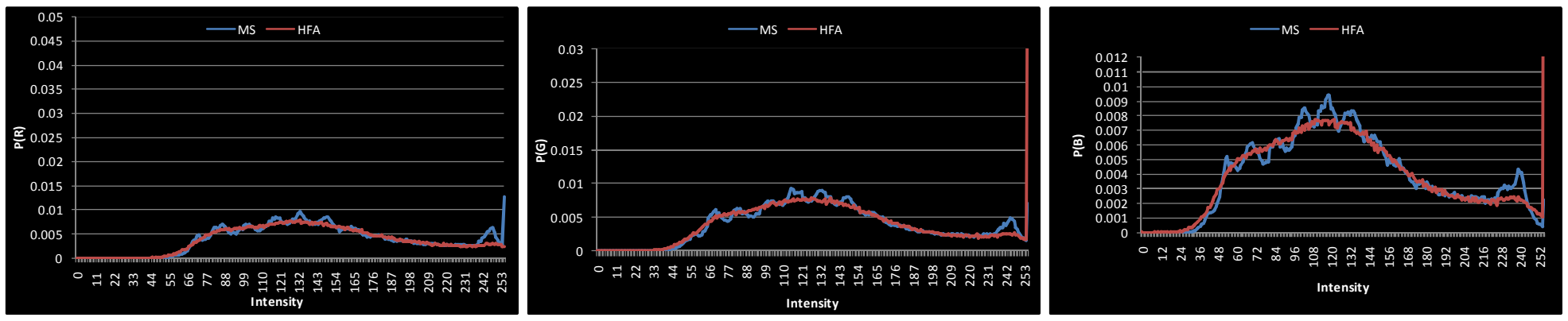

HFA

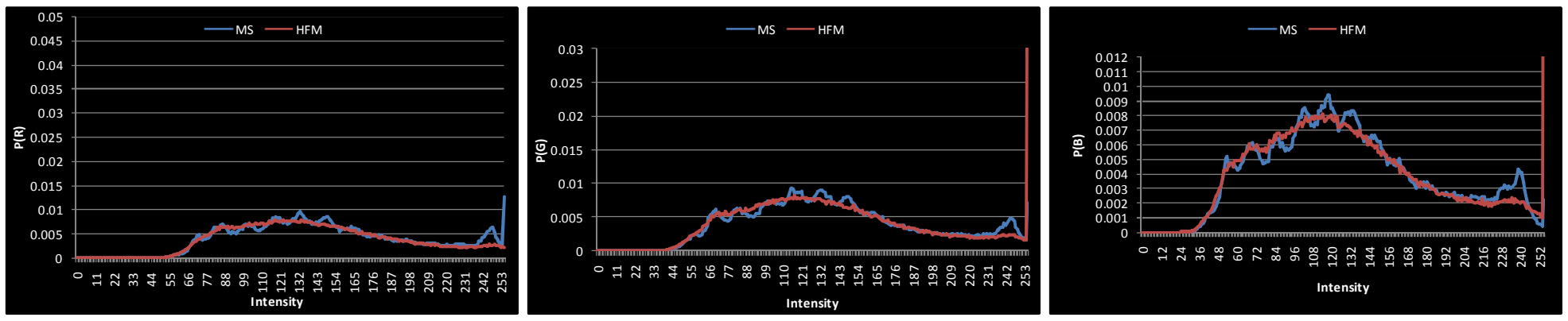

HFM

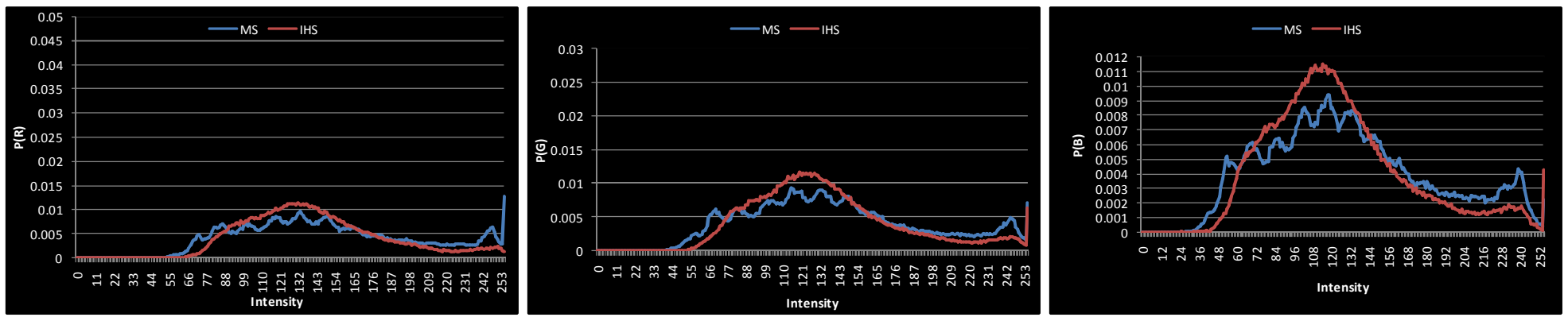

IHS

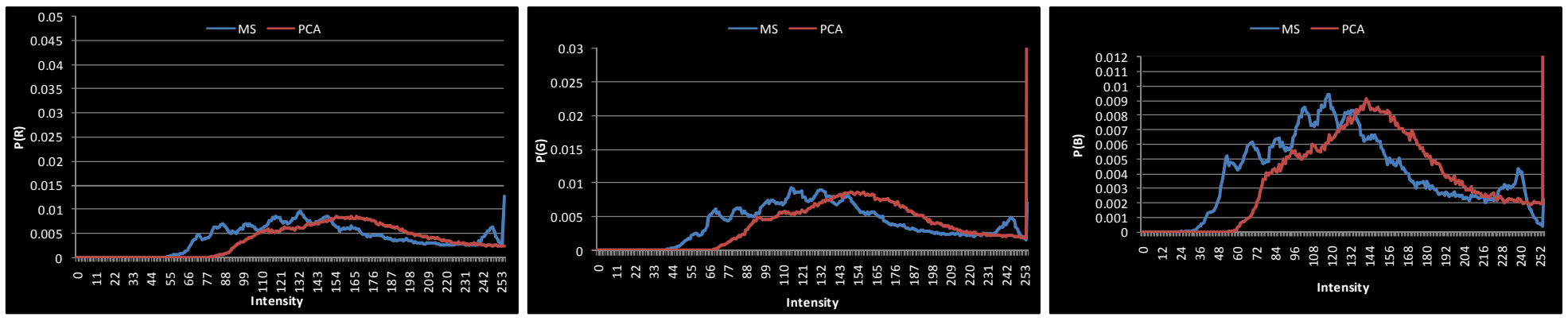

PCA

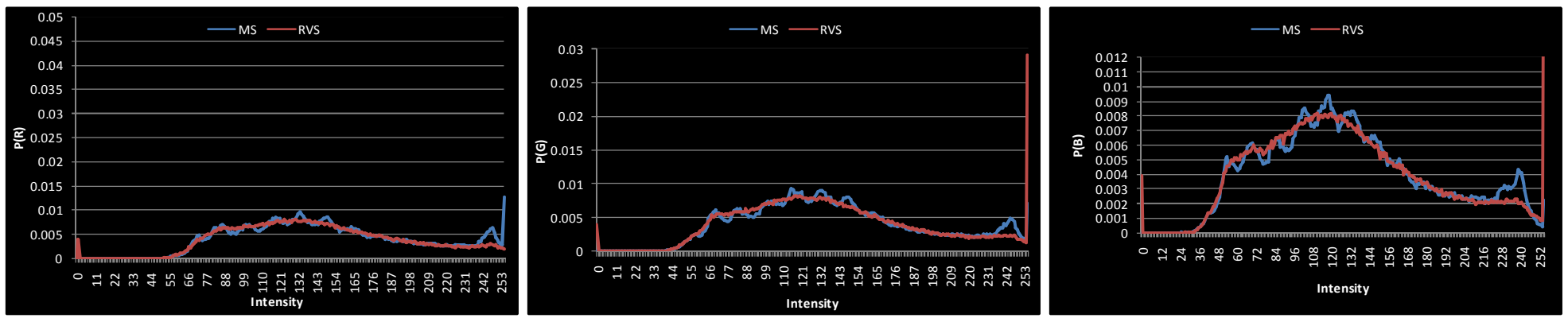

RVS

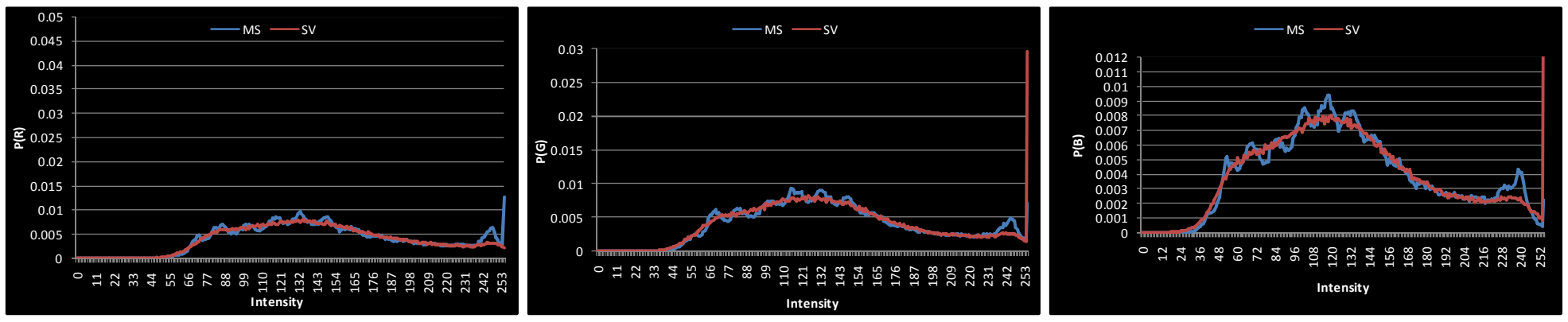

SF

**Fig.15**: Histogram Analysis for All RGB- Color Bands of Completely Fused Images with MS Images

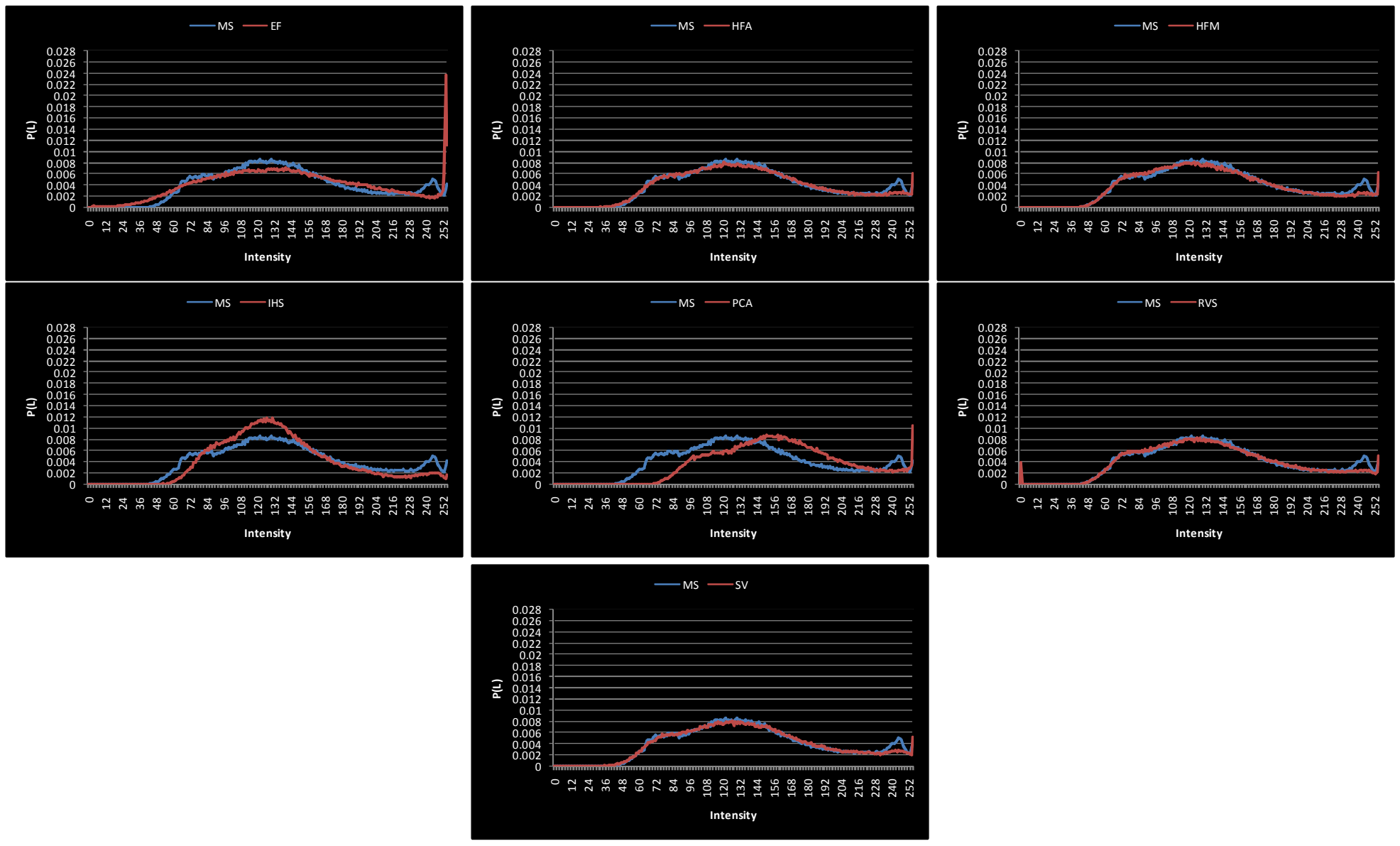

Fig. 16: Histogram Analysis L-component for Whole of the Fused Images with MS Image

## V. CONCLUSION

This study proposed a new measure to test the efficiency of spatial and spectral resolution of the fusion images applied to a number of methods of merge images. These methods have obtained the best results in our previous studies and some of them depend on the pixel level fusion including HFA, HFM, IHS and RVS methods while the other methods based on features level fusion as fallows PCA, EF and SF method. Results of the study show the importance to proposed new CSA as a criterion to measure the quality evaluation for the spatial resolution of the fused images, in which the results showed the effectiveness of high efficiency when compared with the other criterion methods for measurement such as the MTF. The study proved the importance of analysis using the edge, which is more accurate and objective than different one that depending on the selection regions or even the whole image to test the spatial improvement for the fused images. This is because the edges are really showing the improvement of the spatial resolution of the images, whereas there is no apparent spatial improvement in the homogeneous regions. In addition, the edges are more affected than the homogenous regions in the image through the processing of the merge by moving the spatial details to the multispectral MS image and consequently affect on the spectral features and that showed in the image after the merged. Therefore, it is recommended to use the spectral analysis for the whole image to determine the spectral distortions in the images, whereas the use of edge’s analysis image has shown crucial difference.

According to CSA, SNR and Histogram results of the analysis, SF is the best method applied in this study to determine the best method of merging in conservation spectral characteristics for original MS image and adding the maximum possible of spatial details of PAN image to the fused image.

## AUTHOR

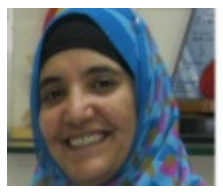

Firouz Abdullah Al-Wassai. Received the B.Sc. degree in, Physics from University of Sana'a, Yemen, Sana'a, in 1993. The M.Sc.degree in, Physics from Bagdad University , Iraqe, in 2003, Research student.Ph.D in the department of computer science (S.R.T.M.U), India, Nanded.

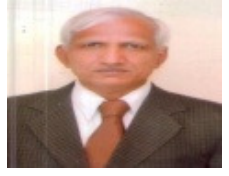

Dr. N.V. Kalyankar, Principal,Yeshwant Mahvidyalaya, Nanded(India) completed M.Sc.(Physics) from Dr. B.A.M.U, Aurangabad. In 1980 he joined as a leturer in department of physics at Yeshwant Mahavidyalaya, Nanded. In 1984 he completed his DHE. He completed his Ph.D. from Dr.B.A.M.U. Aurangabad in 1995. From 2003 he is working as a Principal to till date in Yeshwant Mahavidyalaya, Nanded. He is also research guide for Physics and Computer Science in S.R.T.M.U, Nanded. 03 research students are successfully awarded Ph.D in Computer Science under his guidance. 12 research students are successfully awarded M.Phil in Computer Science under his guidance He is also worked on various boides in S.R.T.M.U, Nanded. He is also worked on various bodies is S.R.T.M.U, Nanded. He also published 34 research papers in various international/national journals. He is peer team member of NAAC (National Assessment and Accreditation Council, India ). He published a book entilteld "DBMS concepts and programming in Foxpro". He also get various educational wards in which "Best Principal" award from S.R.T.M.U, Nanded in 2009 and "Best Teacher" award from Govt. of Maharashtra, India in 2010. He is life member of Indian "Fellowship of Linnean Society of London(F.L.S.)" on 11 National Congress, Kolkata (India). He is also honored with November 2009.

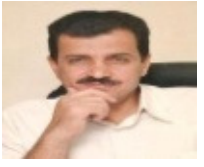

Dr. Ali A. Al-Zuky. B.Sc Physics Mustansiriyah University, Baghdad , Iraq, 1990. M Sc. In1993 and Ph. D. in1998 from University of Baghdad, Iraq. He was supervision for 40 postgraduate students (MSc. & Ph.D.) in different fields (physics, computers and Computer Engineering and Medical Physics). He has More than 60 scientific papers published in scientific journals in several scientific conferences.